\documentclass{article}

\usepackage[final,nonatbib]{compositional_learning}%

\usepackage[utf8]{inputenc} %
\usepackage[T1]{fontenc}    %
\usepackage[USenglish]{babel}

\usepackage{microtype}
\usepackage{amsmath,amsfonts,amssymb,amsthm}

\usepackage{algorithmic}
\usepackage[linesnumbered,ruled]{algorithm2e}
\usepackage[toc,page]{appendix}
\usepackage{bm}
\usepackage{bbm}
\usepackage{booktabs} %
\usepackage{cancel}
\usepackage{caption} \usepackage{subcaption}
\usepackage{centernot}
\usepackage{csquotes}
\usepackage{enumitem}
\usepackage[mathscr]{eucal}
\usepackage{graphicx}
\usepackage{mathtools}
\usepackage{multirow}
\usepackage{nicefrac}       %
\usepackage{tabularx}
\usepackage{thmtools}
\usepackage[table,xcdraw,dvipsnames]{xcolor} 
\usepackage{tcolorbox}
\usepackage{tikz}
\usepackage{dsfont}

\usepackage[colorlinks,citecolor={green!80!black}]{hyperref}
\usepackage{url}

\usepackage[style=numeric,sortcites,maxcitenames=2,maxbibnames=50,natbib,abbreviate=false,useprefix]{biblatex}
\renewbibmacro{in:}{}

\makeatletter %
\AtBeginDocument{\toggletrue{blx@useprefix}}
\AtBeginBibliography{\togglefalse{blx@useprefix}}
\makeatother

\usepackage[capitalize,noabbrev]{cleveref}

\makeatletter
\def\@captype{table}
\makeatother

\newcommand{\bmf}[1]{\bm{\mathsf{#1}}}
\DeclareMathOperator*{\E}{\mathbb{E}}

\newcommand{\R}{\mathbb{R}}

\newcommand{\vG}{\bmf{G}}

\newcommand{\vx}{\bmf{x}}
\newcommand{\vy}{\bmf{y}}

\newcommand{\vz}{\bmf{z}}

\definecolor{cgpt}{HTML}{00E079}
\definecolor{cclaude}{HTML}{9B4400}

\usepackage{todonotes}  %

\title{Understanding Simplicity Bias towards Compositional Mappings via Learning Dynamics}

\author{%
  Yi Ren\\
  University of British Columbia\\
  \texttt{renyi.joshua@gmail.com} \\
   \And
   Danica J. Sutherland \\
   University of British Columbia \& Amii \\
   \texttt{dsuth@cs.ubc.ca} \\
}

\begin{document}

\maketitle

\begin{abstract}
Obtaining compositional mappings is important for the model to generalize well compositionally.
To better understand when and how to encourage the model to learn such mappings,
we study their uniqueness through different perspectives.
Specifically,
we first show that the compositional mappings are the simplest bijections through the lens of coding length (i.e., an upper bound of their Kolmogorov complexity).
This property explains why models having such mappings can generalize well.
We further show that the simplicity bias is usually an intrinsic property of neural network training via gradient descent.
That partially explains why some models spontaneously generalize well when they are trained appropriately.
\end{abstract}

\section{Introduction}
\label{sec:intro}

There is a general belief that having more compositional representations is the key to improving compositional generalization \citep{liu2022towards}.
Although there are many specifically designed algorithms (e.g., \citep{qiu2021improving}) and network structures (e.g., \citep{kuo2020compositional, ren2024improving}) with this goal,
methods to reliably obtain such representations in a variety of settings remain elusive.
On the other hand,
it has been repeatedly shown that compositional generalization ability can spontaneously emerge from standard supervised learning tasks \citep[e.g.][]{power2022grokking} or under repeated self-distillation training \citep[e.g.][]{Ren2020Compositional}.
In a recent position paper,
\citet{huh2024platonic} argued based on a variety of previous results that \textit{deep networks naturally adhere to Occam’s razor, implicitly favoring simple solutions that fit the data}.

To help understand the relationships between compositional mappings and the network's inherent bias,
we first argue that compositional mappings are generally the simplest bijections to learn.
Specifically,
assuming the data-generating process is compositional,
compositional mappings are the simplest
(i.e.\ lowest-complexity).
Next,
we demonstrate experimentally that neural networks naturally favor simpler mappings through the process of gradient descent.
This simplicity bias can be intuitively explained by the mutual influence of learning different samples,
building on prior analyses \citep{ren2022better, guo2024lpntk}.
Although our experiments and discussions are restricted to a simplified setting,
we believe the framework of this paper can help pave the way towards analyzing why some networks naturally achieve great compositional generalization ability under suitable learning tasks,
and hopefully help inspire more effective algorithms to exploit this simplicity bias for more effective generalization.
The code for all experiments can be found in \url{https://github.com/Joshua-Ren/simplicity_bias_learning_dynamics}.

\section{Compositional Mappings are the Simplest Bijections}
\label{sec:comp_K_comp}
In this section,
we first specify that the mapping from the ground-truth characters of the input signal to the learned representation is the key to analyzing the compositional generalization problem.
We then describe the uniqueness of compositional mappings by analyzing their Kolmogorov complexity.
In short, we show that compositional mappings are simpler than non-compositional bijections.

\begin{figure}[t]
    \begin{center}
    \centerline{\includegraphics[width=0.9\textwidth]{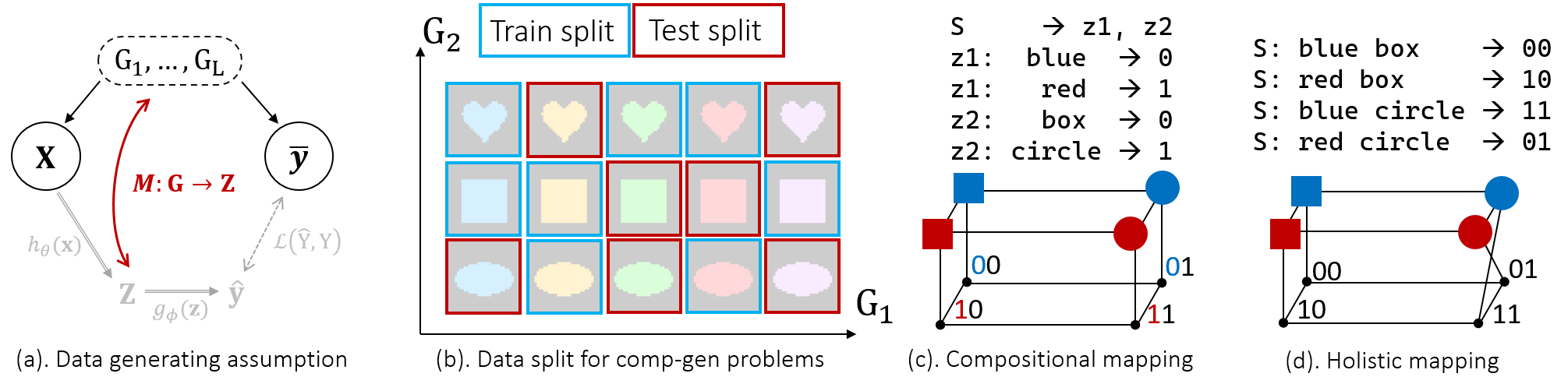}}
    \caption{The compositional generalization problem (\textbf{a, b}) and two types of bijections (\textbf{c, d}).
            }
    \label{fig:problem_setting}
    \end{center}
\vskip -0.2in
\end{figure}

We start with a typical compositional generalization problem,
where the input signal $\vx$ and label(s) $\bar{\vy}$ are determined by several ground-truth generating factors $\vG\triangleq[G_1,\dots, G_L]\in\mathcal{G}$,
as in \cref{fig:problem_setting}-(a).
All $G_i$ are discrete variables with $V$ possible values.\footnote{Continuous $G_i$ can be quantized to fit our analysis; factors with fewer than $V$ values can randomly assign features or pad zeros.}
In this problem,
the model only learns from data samples generated from a subset of all possible $\vG$ and tries to generalize to unseen $\vG$.
Consider a colored-dSprite \citep{dsprites17} example provided in \cref{fig:problem_setting}-(b),
where $G_1$ and $G_2$ are the shape and color of the objects.
With the provided train/test split,
the model must first learn the concepts of ``red'' and ``box'' separately,
and then compositionally combine them to make a correct prediction on the ``red box'' in the test set.
Suppose the model uses a deep neural network $h_\theta:\mathcal{X}\rightarrow\mathcal{Z}$ to extract features from the input signals,
where the representation space $\mathcal{Z}$ is also an $L\times V$ discrete space.
Assuming each $\vx$ has a unique $\vG$,
we can then define the mapping between ground-truth factors and the extracted representations as $M:\mathcal{G}\rightarrow\mathcal{Z}$.
In order to generalize well,
a good mapping $M$ should satisfy:
\begin{itemize}
    \item $M$ should be a bijection, otherwise, two $\vx$ with different $\vG$ will be mapped to the same $\vz$, 
           which makes it impossible for the task head to separate these two $\vx$ from their $\vz$;
    \item $M$ should ensure different $z_i$ consistently encode different $G_j$ in a non-overlapping way, so the model can generalize appropriately to novel combinations.
\end{itemize}

To get a clearer picture of this second requirement,
we consider a ``Toy256'' example,
where $G_1=\{\texttt{blue,red}\}$, $G_2=\{\texttt{box,circle}\}$, and
$\vz=\{00, 01, 10, 11\}$.
There are $4^4=256$ possible mappings for $M$,
of which $4!=24$ are bijections.
Among all these bijections,
only $2! \times 2^2=8$ are compositional.
To get such a mapping,
we should first assign color and shape to different $z_i$,
and then separately assign $0$ or $1$ to represent different values for each attribute.
Following \citet{Ren2020Compositional},
we call those non-compositional bijections ``holistic mappings''.
As demonstrated in \cref{fig:problem_setting}-(c, d),
compositional mappings can be decomposed into some shared rules while holistic mappings cannot,
as illustrated by corresponding CFGs (context-free grammars) at the top of the figure.

Since the compositional mappings are generated in a systematic way,
intuitively they are simpler and can be described with less effort.
We can use group theory to define the simplicity of a bijection more formally.
When generating a compositional mapping,
we first select $z_i$ for each $G_j$ in a non-overlapping way.
Such a process can be represented by an element in a symmetry group $S_L$.
We then build an injection from the paired $G_j$ to $z_i$,
by which each possible value of the $j$-th attribute is encoded by different ``words'' in $z_i$.
In short,
assuming both $\vG$ and $\vz$ are $L\times V$ grid spaces,
any compositional mapping can be described by an element in the group $S^L_V\rtimes S_V\in S_{V^L}$,
where $\rtimes$ is the semidirect product in group theory.

This implies why a compositional mapping has a lower Kolmogorov complexity upper bound\footnote{While it is not possible to lower-bound the Kolmogorov complexity of any particular mapping without fixing the underlying Turing machine, a counting argument shows that \emph{most} non-compositional bijections must have higher complexity.} than an arbitrary non-compositional one among all bijections.
From the definition of the symmetry group,
we know each element in $S_{V^L}$ can be represented by a permutation matrix of size $V^L$.
As there is only one ``$1$'' in each row and column of a permutation matrix,
any permutation matrix can be uniquely represented by a permuted sequence of length $V^L$.
Specifically, assume we have a sequence of natural numbers $\{1,2,...,V^L\}$,
each permuted sequence $\mathsf{Perm}(\{1,2,...,V^L\})$ represents a distinct permutation matrix,
and hence represents a distinct bijection from $\vG$ to $\vz$.
In other words,
we can encode one bijection from $\vG$ to $\vz$ using a sequence of length $V^L$, 
i.e., $\mathsf{Perm}(\{1,2,...,V^L\})$,
and bound the corresponding Kolmogorov complexity (in bits) as
\begin{equation}
    \mathcal{K}(\text{bijection}) \le V^L\cdot \log_2 V^L=V^L\cdot L\cdot \log_2 V,
    \label{eq:app_KC_bijection}
\end{equation}
As an arbitrary bijection from $\vG$ to $\vz$ doesn't have any extra information to improve the coding efficiency,
\cref{eq:app_KC_bijection} provides an upper bound of the minimal Kolmogorov complexity.

On the contrary,
as each compositional mapping can be represented by an element in $S_V^L\rtimes S_L$,
we can encode the mappings more efficiently.
Specifically, we need to first use $L$ sequences with length $V$,
i.e., $\mathsf{Perm}(\{1,2,...,V\})$,
to represent the assignment of ``words'' for each $z_i$.
After that, we need one sequence of length $L$,
i.e., $\mathsf{Perm}(\{1,2,...,L\})$ to encode the assignment between $z_i$ and $G_j$.
Ignoring the necessary separators for these $L+1$ sub-sequences,
the corresponding Kolmogorov complexity is then bounded as
\begin{equation}
    \mathcal{K}(\text{comp}) \le V\cdot\log_2 V + L\cdot\log_2 L.
    \label{eq:app_KC_comp}
\end{equation}    

To compare the Kolmogorov complexity of different mappings,
we can define a ratio as $\gamma\triangleq \frac{\mathcal{K}(\text{bijection})}{\mathcal{K}(\text{comp})}$.
Obviously, when $L\leq V$,
$\gamma\geq\frac{V^{L-1}\cdot L}{2}$,
which is larger than 1 as long as $L,V\geq 2$.
When $L>V$,
$\gamma\geq\frac{V^L\log_2 V}{2\log_2 L}$,
which is also larger than 1 when $L,V\geq 2$.

Note that there might be some partially compositional mappings.
For example, we can have a mapping with $z_{i\leq 10}$ sharing the reused rules while other $z_{i>10}$ doesn't.
Then this type of mapping can be represented by an element in $S_V^{10}\rtimes S_{10}\rtimes S_{V^{L-10}}$.
As a mapping in this subset shares 10 common rules,
its Kolmogorov complexity is between $\mathcal{K}(\text{bijection})$ and $\mathcal{K}(\text{comp})$.
Intuitively,
\emph{for all bijections},
smaller $\mathcal{K}(\cdot)$ means higher compositionality.

\section{Simpler Mappings are Learned Faster}
\label{sec:faster}
The analysis above links the concepts of compositionally, simplicity, and Kolmogorov complexity under an idealized setting,
which also aligns well with many related works.
For example,
\citet{huh2024platonic}
claim that simplicity bias is the key for the models in different modalities converging to a shared representation space that is similar to the ground truth.
\citet{goldblum2023no} also link Kolmogorov complexity to PAC-Bayes generalization bounds.
This supports the idea that having more compositional mappings greatly benefits the model's generalization ability.
This section further demonstrates that a neural network naturally favors such simpler mappings.
We will first verify this claim by experiments under manual settings,
and then provide a detailed explanation of why such a tendency exists using learning dynamics.

Specifically,
we claim that simpler mappings are learned faster by a neural network trained using GD.
To verify this,
we consider a multi-label classification problem and create 256 different datasets (each only contains 4 examples) for each $M$ in our ``Toy256'' setting.
For example,
the dataset for the mapping in \cref{fig:problem_setting}-(c) should be \{(\texttt{blue box}, 00), (\texttt{blue circle}, 01), (\texttt{red box}, 10), (\texttt{red circle}, 11)\},
where the label ``01'' means $\bar{y}_1=0$ and $\bar{y}_2=1$.
We then randomly initialize a neural network as our $h_\theta$ and concatenate two $\mathsf{Sigmoid}$ functions as our $g_\phi$.
With the same initialization and all hyper-parameters,
we train the network to convergence for each dataset.
We also consider different input signals (images and dense random vectors),
network structures (MLP and ResNet),
loss functions (cross-entropy and mean square error),
and optimizers (standard SGD and Adam).
Please refer to \cref{app:sec:settings} for more details.

\begin{figure}[t]
    \begin{center}
    \centerline{\includegraphics[width=0.9\textwidth,trim=0 0 0 0, clip]{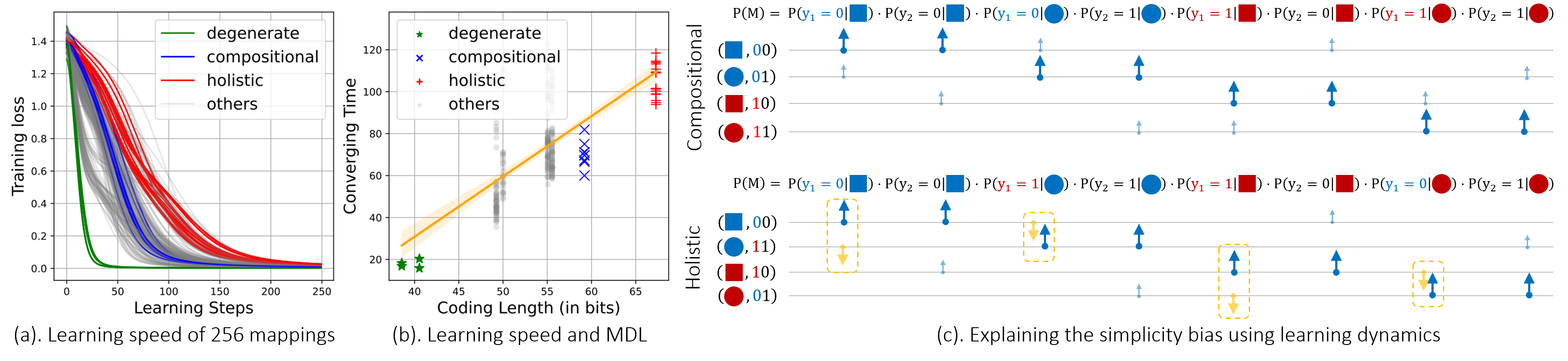}}
    \caption{The evidence and explanations of the claim that \textit{simpler mappings are learned faster}.
             The blue arrows in the last panel mean when learning the given example,
             the model increases its confidence in the corresponding prediction.
             The increase of the darker arrows is stronger than that of lighter ones
             because the confidence changes for the lighter ones are indirectly caused by the ``elasticity'' of the neural network \citep{He2020elasticity}.
             For holistic mapping,
             the yellow arrows pointing down mean the corresponding confidence decreases after learning the example.
            }
    \label{fig:toy256}
    \end{center}
\vskip -0.2in
\end{figure}

\cref{fig:toy256}-(a) shows the training curves of 256 runs in our default setting.
Since the only difference among these runs is the dataset generated by different mappings,
it is safe to conclude that the difference in their learning speed is caused by \textit{the inherent bias of the model's learning behavior} on this problem.
From this figure,
we see compositional mappings are learned faster than holistic ones.
However,
some mappings are learned even faster,
which makes sense because those non-bijection mappings contain degenerate components,
i.e., two or more objects are mapped to the same $\vz$,
which means they are simpler.
That also explains why the four degenerate mappings,
which map all four objects to the same $\vz$,
are learned fastest among all 256 mappings.

To further verify this,
we quantify the learning speed using the concept of ``convergence time,''
i.e., the area under the learning curve.
A smaller convergence time means the mapping is learned faster.
This metric is similar to the C-score of \citet{c_score},
which describes a training example's difficulty.
Also,
if the model is trained with cross-entropy loss and all examples only appear once,
this metric is the compression rate for the entire dataset \citep{compression_for_AGI}.
These works also hint that learning speed is deeply related to compression, simplicity, and generalization ability.

Another quantity we want to explicitly calculate is bounds on each mapping's Kolmogorov complexity.
Since the mapping space studied in our Toy256 setting is simple enough,
we can first create the CFGs for each mapping and then convert them to a piece of description sequence using the method provided by \citet{kirby2015compression}.
After that,
we can use Huffman coding \citep{cormen2022introduction} and calculate the coding length in bits for each mapping.
Please refer to \cref{app:sec:coding_length} for more details.
In short,
a smaller coding length means the mapping is simpler.
It is clear in \cref{fig:toy256}-(b) that compositional mappings are the simplest bijections.
Note that all the non-bijection mappings contain degenerate components,
hence are simpler than bijections.
The figure also clearly demonstrates that the learning speed is strongly correlated to the coding length (with $\rho>0.65$ and $p<10^{-30}$),
matching our hypothesis well.
This trend is consistent across various settings,
as demonstrated in \cref{app:sec:more_results}.

The results above bridge the simplicity bias to the model's learning behavior,
where the latter can be further explained using learning dynamics \citep{ren2022better, ren2023how}.
Remember our model generates probabilistic predictions on both $y_1$ and $y_2$ using $\mathsf{Sigmoid}$ functions.
Then, we can directly write down the predicted probability of each mapping as a product of eight terms,
as in the top line of \cref{fig:toy256}-(c).
In this figure,
we demonstrate how the model's confidence of different $M$ is updated when learning specific training samples.
For example, in the first row of the compositional mapping in the figure,
the model learns $(\texttt{blue box}, 00)$.
Then the corresponding $P(y_1=0\mid\texttt{blue box})$ and $P(y_2=0\mid\texttt{blue box})$ are significantly improved,
since they are directly updated by learning this example.
Furthermore,
as the neural network has local elasticity \citep{He2020elasticity},
the model's predictions on those ``similar'' (measured using Hamming distance) input examples would also be indirectly updated (represented by the small arrows in the figure).
As a result,
the model's confidence on \texttt{blue circle} and \texttt{red box},
which share one attribute with the learned \texttt{blue box},
are influenced more by this update.
Furthermore,
since a compositional mapping always utilizes consistent values to represent the same attribute (e.g., $z_1=0$ always encodes the blue color),
all the direct and indirect updates align well with the compositional mapping.
That is why the training loss of such mappings decreases faster.
On the contrary,
for a holistic mapping,
we observe several contradictions between the direct and indirect updates:
learning it requires the model to use more updates to counteract those negative indirect influences.
That explains why compositional mappings are usually learned faster than holistic ones by a neural network.

\section{Conlusion}
\label{sec:conclusion}
This paper first shows that compositional mappings are the simplest bijections in terms of Kolmogorov complexity or coding length.
Then, using experiments and analysis from the learning dynamics perspective,
the paper claims that the simplicity bias (in terms of coding length) is inherent in neural network training for typical architectures using gradient descent.
Although the settings in the paper are simple,
the theoretical formulation and analysis have the potential to be extended to more practical problems.

\printbibliography

\clearpage

\clearpage

\newwrite\pageout 
\immediate\openout\pageout=splitpage.txt\relax
\immediate\write\pageout{\thepage}
\immediate\closeout\pageout

\clearpage
\appendix

\section{Compositional Representation and Platonic Representation Hypothesis}
\label{app:sec:mdl_topsim}
This appendix tries to uncover the implicit relationship between compositional representation learning (usually studied in a manually toyish setting) and the Platonic representation hypothesis (proposed in \citep{huh2024platonic}, experimentally verified on many SOTA large vision and language models).
Specifically,
we focus on the following three aspects:
1.) the underlying assumption of the existence of $\vG$;
2.) the measuring metrics;
3.) the converging pressures.
Our analysis hints that more advanced compositional generalization ability could also be achieved if we design appropriate learning systems following the fundamental principles demonstrated in \citet{huh2024platonic}.

\subsection{The Underlying Assumption of the Ground-truth Generating Mechanism}
The main claim of \citet{huh2024platonic} is that there exists a unique ground-truth idealized world (i.e., the $\vG$ in our paper),
from which, all observations in different modalities are its projections.
A deep learning system,
which learns from these projections and then generalizes to related tasks,
are trying to uncover such ground truth.
As the models in different modalities become stronger,
their representations (i.e., $\vz\in\mathcal{Z}$ in our paper) are more aligned,
because they all tend to converge to the ground truth $\vG$.

In our paper,
we also assume the existence of $\vG$ and consider both input signal $\vx$ and labels $\bar{y}$ are determined by it.
By treating the mapping from $\mathcal{G}\rightarrow\mathcal{Y}$ as a special projection for a specific modality,
our \cref{fig:problem_setting}-(a) becomes the upper part of the Figure 1 in \citep{huh2024platonic}.
The goal of a compositional representation learning task is to recover a good representation space that is similar to the ground truth,
and hence generalize well to unseen combinations of attributes.
This also aligns with the claim that ``models generalize better on different modalities align better to the ground-truth'' in \citep{huh2024platonic}.

\subsection{Measuring Metrics: Kernel Alignment, Disentanglement, and Topological Similarity}
To mathematically describe the representation's convergence,
\citet{huh2024platonic} use three steps to define a metric called \textit{Kernel Alignment} to measure the similarity between two representation spaces.
\begin{itemize}
    \item[1. ] A \textit{representation}, which maps the input signal to a dense representation space, is a function $f:\mathcal{X}\rightarrow \mathbb{R}^n$.
                Note that our $h_\theta(\vx)$ plays a similar role;
    \item[2. ] A \textit{kernel}, $K:\mathcal{X}\times\mathcal{X}\rightarrow\mathbb{R}$, characterize the similarity between two elements in $\mathcal{X}$.
                In a dense representation case, the inner product is usually applied, i.e., $K(x_i,x_j)=\langle f(x_i),f(x_j) \rangle, K\in\mathcal{K}$.
                Our paper consider Hamming distance, because our $\vG$ and $\vz$ are all categorical variables;
    \item[3. ] A \textit{kernel-alignment metric} $m:\mathcal{K}\times \mathcal{K}\rightarrow\mathbb{R}$, measures the similarity between two kernels.
\end{itemize}

For the third-level measurement, \citet{huh2024platonic} use Centered Kernel Distance (CKA),
a kernel-alignment metric throughout their paper.
Actually,
many related works in compositional generalization also have similar measurements,
e.g., the topological similarity proposed in \citep{brighton2006understanding}:
\begin{equation}
    \mathsf{Topsim}(\vG, \vz)\triangleq\mathsf{Corr}\left(d_G(\vG^{(i)},\vG^{(j)}), d_z(\vz^{(i)},\vz^{(j)}) \right),%
    \label{eq:topsim}
\end{equation}
This definition also follows three steps:
$\vz$ are representations generated by feeding $\vx$ to $h_\theta$,
$d_G$ and $d_z$ are distance measurements (or kernel in the second step above) for space $\mathcal{G}$ and $\mathcal{Z}$,
and $\mathsf{Corr}(\cdot,\cdot)$ is the Spearman's correlation measuring the relationship between the output of two functions (kernels).
In short,
\textit{higher topological similarity means similar objects in $\mathcal{G}$ are mapped to similar positions in $\mathcal{Z}$.}
If we consider $\mathcal{G}$ as another modality of the projected ground truth,
the topological similarity is just a special kernel-alignment metric used in \citep{huh2024platonic}.
Also, some other measurements of compositionality like TRE (Tree Reconstruction Error, \citep{andreas2018measuring}),
representation disentanglement \citep{higgins2018towards}, etc., also follow this principle generally.
In summary,
since the main measurement of the Platonic representation hypothesis and the compositional representation learning are essentially identical,
we can draw more parallels between these two seemingly distinct fields in the future.

\subsection{The Converging Pressures}
Section 3 of \citep{huh2024platonic} proposes three pressures that lead the model's representations to converge to the ground truth.
We could also find some counterparts in the field of compositional representation learning.
The first one is task generality,
which means requiring the model to solve more tasks using the same representations leads to better convergence to the ground truth.
We can also draw a similar conclusion from the experiments in \citep{ren2024improving},
in which the authors show that the quantity and diversity of the learning tasks play an important role in achieving good systematic representations.

The second pressure is the model's capacity. Because bigger models are more likely to converge to a shared representation than smaller ones \citep{huh2024platonic}.
This partially aligns with our ``unambiguous'' requirement on the mapping $M:\mathcal{G}\rightarrow\mathcal{Z}$ discussed in \cref{sec:comp_K_comp}.
This requires the model to be capable enough to have perfect training accuracy,
otherwise, important information about the task labels would be lost.
However,
\citet{zhang2021understanding} demonstrates that achieving perfect training accuracy (even if the label is purely random noise) is not a hard task for a deep neural network, while the optimal mapping can be even simpler than the noisy-label dataset.
As a result,
it might be more useful to consider the influence of the model's capacity from a dynamical perspective (e.g., optimization),
which is usually neglected when studying the relationship between generalization and model capacity.

The last pressure, i.e., the simplicity bias, is the one we discussed most in our paper.
We showed that such simplicity can be understood as the lower bound of Kolmogorov complexity,
which also measures how compressible the mapping is.
Requiring simpler mappings also aligns well with Occam's Razor,
which might be an important direction for our future exploration.
The explanation of learning dynamics is a good starting point for combining this model-agnostic measurement (i.e., Kolmogorov complexity) with the model's inherent bias.
Actually,
Figure 1 of \citet{zhang2021understanding} shows that under the same setting,
the model learns the random noise labels slower than the ground truth labels.
This phenomenon could also be explained by the interactions between training examples used in this paper.
Since the random noisy-label dataset would have two semantically similar $\vx$ with very different labels,
the direct and indirect updates would have more contradictions,
similar to the holistic updating case in \cref{fig:toy256}.

\section{Coding Length and Topological Similarity for the Mappings in Toy256}
\label{app:sec:coding_length}

\begin{figure}[h]
    \begin{center}
    \centerline{\includegraphics[width=0.8\textwidth,trim=0 0 0 0, clip]{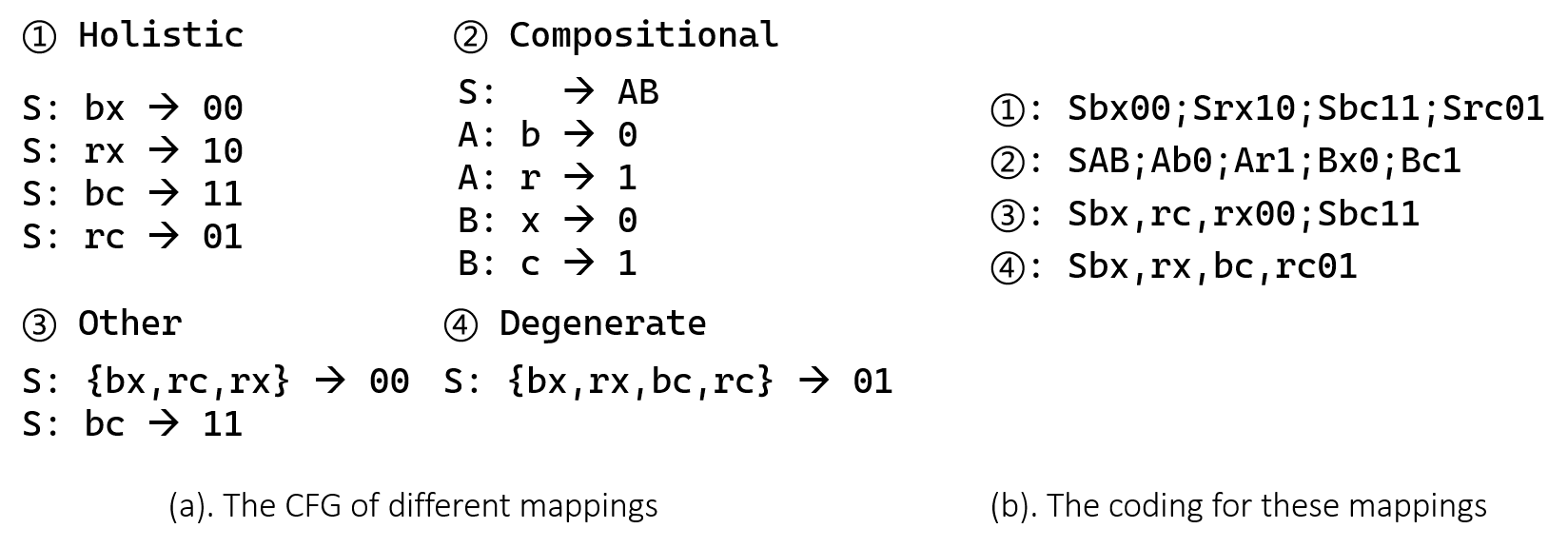}}
    \caption{Four typical mappings studied in this paper and their coding strings.}
    \label{app:fig:coding_length}
    \end{center}
\vskip -0.2in
\end{figure}

The main target of this paper is to show that the simplicity bias is inherent in neural network training.
Inspired by many related works on compositional generalization,
we believe the Kolmogorov complexity is a perfect measurement for the simplicity of a mapping.
However, it is well known that Kolmogorov complexity is usually hard to calculate and people typically use the minimum description length (MDL) under specific constraints as its approximation \citep{mackay2003information}.

To experimentally show the correlation between learning speed and simplicity,
we use the coding method provided in \citet{kirby2015compression} to calculate the coding length for all 256 mappings.
Note that such a coding mechanism might not be optimal (i.e., its length is not the MDL).
Hence we only call this measurement coding length (CL) throughout the paper.

The calculation of CL involves three steps.
First,
we convert all mappings using a \textit{compressed} CFG,
as illustrated in \cref{app:fig:coding_length}-(a).
As all the attributes studied here are categorical variables,
we use \texttt{b, x, r, c} to represent \texttt{blue, box, red, circle}, respectively.
Then,
we delete all the redundant characters and generate the unique coding sequence for each CFG,
as in \cref{app:fig:coding_length}-(b).
The special characters ``\texttt{S}'' and ``\texttt{;}'' denote the starting and ending of one piece of rule and ``\texttt{,}'' is used to separate different objects sharing the same message.
Finally,
the coding length in bits of a mapping $M$ is calculated using
\begin{equation}
    \mathsf{CL}(M) = -\sum_{i=1}^{|\mathsf{Seq}(M)|} \log_2 p(s_i);\quad p(s_i)=\frac{\mathsf{Cnt}(s_i)}{|\mathsf{Seq}(M)|}
    \label{eq:app:CL}
\end{equation}
where $\mathsf{Seq}(M)$ converts the mapping to a coding sequence and $p(s_i)$ is the probability of the $i$-th character in this code squence.
For example, for the degenerate sequence \texttt{[Sbx,rx,bc,rc01]},
$p(s_1)=p(s_7)=\frac{\mathsf{Cnt}(\texttt{b})}{|\mathsf{Seq}(M)|}=\frac{2}{14}$.

Furthermore,
we also verify the correlation between the learning speed and topological similarity defined in \cref{eq:topsim}.
Generally,
the $\mathsf{Topsim}$ for perfect compositional mappings equals one.
Note that the $\mathsf{Topsim}$ of pure degenerate mappings that maps every object to the same $\vz$ is not well defined in \cref{eq:topsim},
because the Spearman's correlation is calculated by $r_s=\frac{\mathsf{Cov}(R[\vz], R[\vG])}{\sigma_{R[\vz]}\sigma_{R[\vG]}}$.
In a degenerate mapping,
the six pair-wise distances of all four possible $\vz$ are all zeros,
which means $R[\vz]=\texttt{[0,0,0,0,0,0]}$.
Hence the corresponding $r_s$ becomes $\frac{0}{0}$.
However,
following the definition of topological similarity that high $\mathsf{Topsim}$ mappings tend to assign similar $\vz$ to $\vx$ with similar $\vG$,
we just define the $\mathsf{Topsim}$ of those degenerate mappings as one.

\section{Experimental Settings}
\label{app:sec:settings}

We consider various settings for the Toy256 examples to verify that the simplicity bias discussed in this paper is general enough.
For the input signals,
we first consider two types of one-hot concatenation vectors.
One is the concatenation of two $2$-dim vectors (\texttt{OHT2} for short).
For example, \texttt{blue box} and \texttt{red circle} are encoded as $\texttt{[0101]}\cdot W_{4\times d}$ and $\texttt{[1010]}\cdot W_{4\times d}$, respectively,
where $W_{4\times d}$ is a randomly initialized matrix fixed for all 256 mappings.
Another setting is \texttt{OHT3},
which considers a redundant dimension for each attribute,
where \texttt{blue box} and \texttt{red circle} are encoded as $\texttt{[010010]}\cdot W_{6\times d}$ and $\texttt{[100100]}\cdot W_{6\times d}$, respectively.
We also consider the vision input,
where the image is sampled and colored from the dSprite dataset,
as illustrated in \cref{fig:problem_setting}.

We consider different network structures for different input modalities.
For the one-hot input,
we use an MLP with three hidden layers with a width of 128.
For the image input,
we consider both a 3-layer MLP and a ResNet9 \citep{he2016deep} with narrower hidden layers.
The task heads for all the networks are identical:
we add two separate linear projection layers with size $h\times2$ on the output of the backbone.
After that,
we take $\mathsf{Softmax}$ on each of these outputs to generate probabilistic predictions.
When calculating the loss function,
we consider both cross-entropy (CE) loss and a mean square error (L2) loss,
where the latter is calculated between the predicting probability vector and a one-hot distribution of the ground truth labels.
When optimizing the network,
we consider both stochastic gradient descent (SGD) and Adam.
Unless otherwise stated,
the learning rate is set to $10^{-3}$, weight decay is $5*10^{-4}$,
and all other parameters are set to be the default values.
Note that all hyper-parameters (including the initialization of the network) are shared for all 256 experiments in each group.

\section{More Experimental Results}
\label{app:sec:more_results}
\begin{figure}[t]
    \begin{center}
    \centerline{\includegraphics[width=1\textwidth,trim=0 0 0 0, clip]{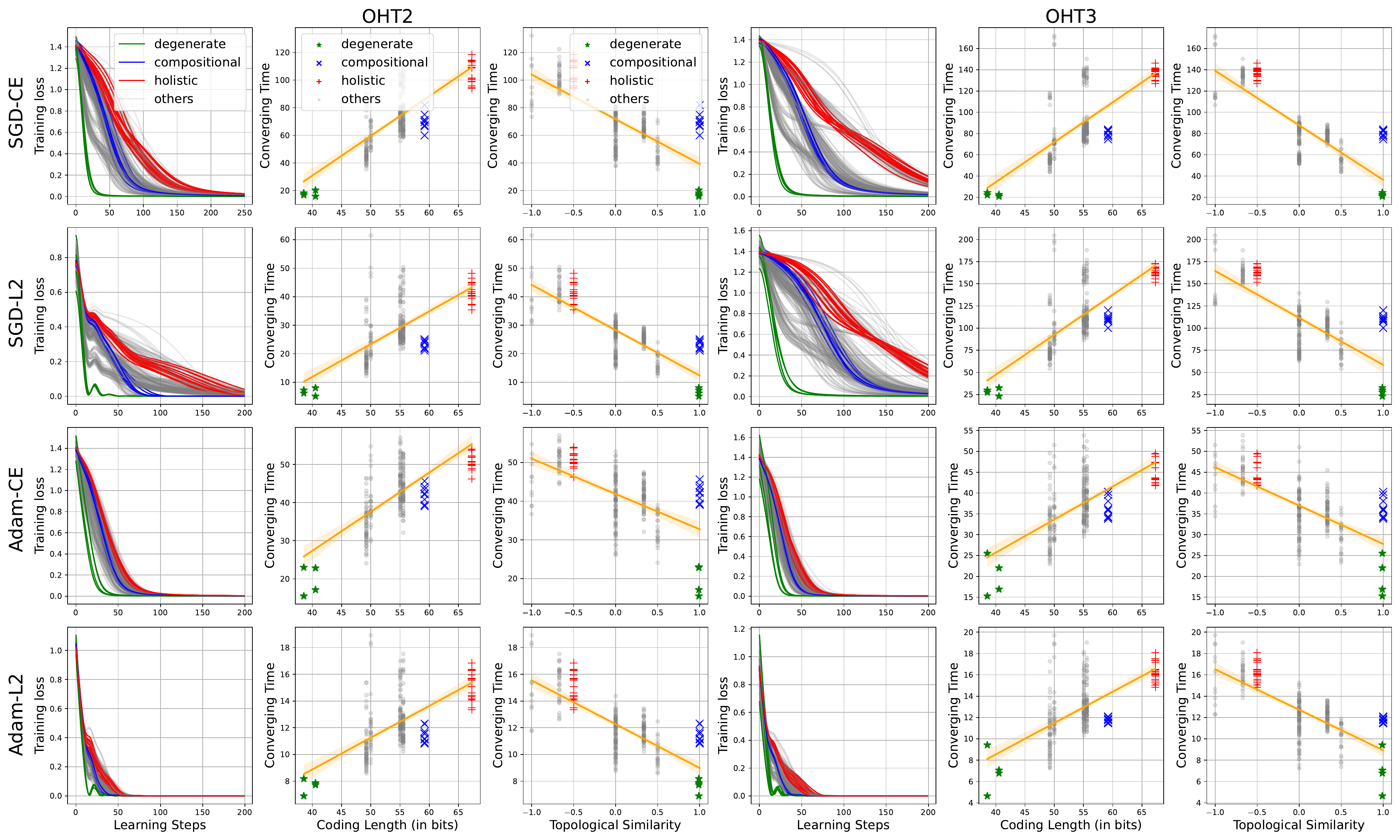}}
    \caption{Experiments for the one-hot vector inputs.
             }
    \label{app:fig:rho_p:oht}
    \end{center}
\vskip -0.2in
\end{figure}

\begin{figure}[t]
    \begin{center}
    \centerline{\includegraphics[width=1\textwidth,trim=0 0 0 0, clip]{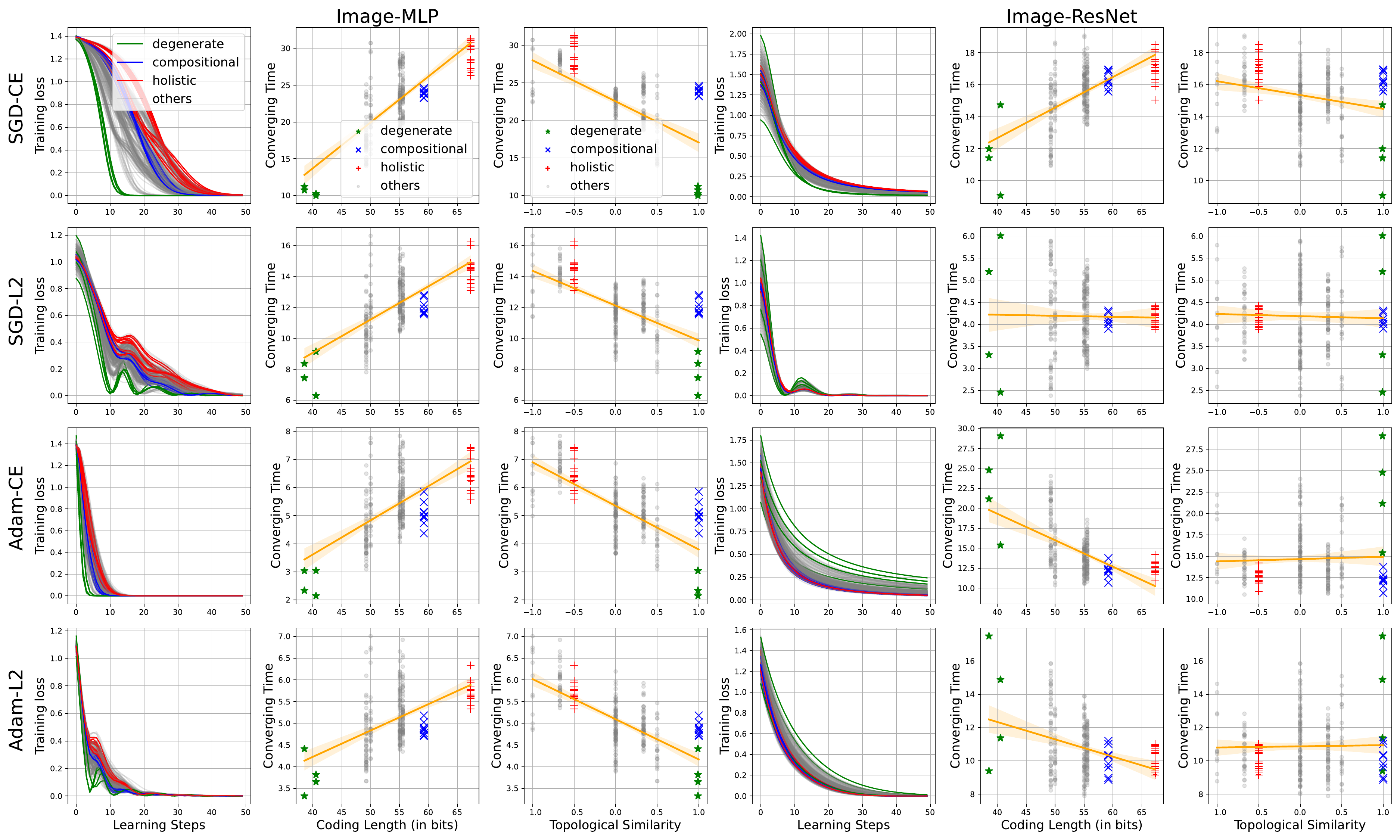}}
    \caption{Experiments for the vision inputs.
             }
    \label{app:fig:rho_p:img}
    \end{center}
\vskip -0.2in
\end{figure}

To visualize the relationship between learning speed and the simplicity of each mapping,
we provide three types of visualizations in \cref{app:fig:rho_p:oht} and \cref{app:fig:rho_p:img}.
The first one is the learning curves of all 256 mappings.
It is clear that under most settings,
the blue curves (i.e., those for compositional mappings) decay faster than the red ones (the non-compositional bijections).
The second one is the scatter plots showing the correlation between the converging time (i.e., the integral under the learning curve) and CL in \cref{eq:app:CL}.
The third one is the scatter plots showing the correlation between the converging time and topological similarity defined in \cref{eq:topsim}.
We also calculate the Pearson correlation in the latter two cases in \cref{app:tab:rho_p}.
It is clear that in most cases,
simpler mappings are indeed learned faster under different settings.
One exceptional case is training a ResNet with image input using Adam optimizer.
The simplicity bias is even reversed compared with the results using SGD.
This phenomenon hints that the bias in DNN's learning is also influenced by the inherent bias of specific network structures and optimizers.
We left this for our future work.

\begin{table*}[h]
  \centering
  \caption{The statistical correlation between learning speed and simplicity.}
  \vskip -0.1in
  \resizebox{\textwidth}{!}{

\begin{tabular}{ccccccc|ccccccc}
\hline
                                                                              &                                 &                        & \multicolumn{2}{c}{CL-Conv.Time} & \multicolumn{2}{c|}{Topsim-Conv.Time} &                                                                                   &                                 &                        & \multicolumn{2}{c}{CL-Conv.Time}                                                   & \multicolumn{2}{c}{Topsim-Conv.Time}                           \\ \cline{4-7} \cline{11-14} 
\multirow{-2}{*}{Input}                                                       & \multirow{-2}{*}{Optim.}        & \multirow{-2}{*}{Loss} & $\rho$         & $p$              & $\rho$           & $p$                & \multirow{-2}{*}{Input}                                                           & \multirow{-2}{*}{Optim.}        & \multirow{-2}{*}{Loss} & $\rho$                                  & $p$                                       & $\rho$                         & $p$                           \\ \hline
                                                                              &                                 & CE                     & 0.6475         & 8.1*1e-32        & -0.7101          & 1.4*1e-40          &                                                                                   &                                 & CE                     & 0.6866                                  & 5.0*1e-37                                 & -0.5911                        & 1.6*1e-25                     \\
                                                                              & \multirow{-2}{*}{\textbf{SGD}}  & L2                     & 0.5793         & 2.4*1e-24        & -0.7817          & 5.3*1e-54          &                                                                                   & \multirow{-2}{*}{\textbf{SGD}}  & L2                     & 0.5932                                  & 1.1*1e-25                                 & -0.6057                        & 5.1*1e-27                     \\
                                                                              &                                 & CE                     & 0.6598         & 2.3*1e-33        & -0.5731          & 9.4*1e-24          &                                                                                   &                                 & CE                     & 0.5403                                  & 8.4*1e-21                                 & -0.6720                        & 5.5*1e-35                     \\
\multirow{-4}{*}{\textbf{\begin{tabular}[c]{@{}c@{}}OHT2\\ MLP\end{tabular}}} & \multirow{-2}{*}{\textbf{Adam}} & L2                     & 0.5378         & 1.4*1e-20        & -0.7223          & 1.2*1e-42          & \multirow{-4}{*}{\textbf{\begin{tabular}[c]{@{}c@{}}Image\\ MLP\end{tabular}}}    & \multirow{-2}{*}{\textbf{Adam}} & L2                     & 0.4433                                  & 9.5*1e-14                                 & -0.6585                        & 3.4*1e-33                     \\ \hline
                                                                              &                                 & CE                     & 0.5976         & 3.5*1e-26        & -0.7963          & 2.2*1e-57          &                                                                                   &                                 & CE                     & 0.6711                                  & 7.4*1e-35                                 & -0.2619                        & 2.2*1e-5                      \\
                                                                              & \multirow{-2}{*}{\textbf{SGD}}  & L2                     & 0.6386         & 9.8*1e-31        & -0.7311          & 4.5*1e-44          &                                                                                   & \multirow{-2}{*}{\textbf{SGD}}  & L2                     & {\color[HTML]{C0C0C0} -0.015}           & {\color[HTML]{C0C0C0} 0.8159}             & {\color[HTML]{C0C0C0} -0.0297} & {\color[HTML]{C0C0C0} 0.6358} \\
                                                                              &                                 & CE                     & 0.5672         & 3.4*1e-23        & -0.6418          & 4.1*1e-31          &                                                                                   &                                 & CE                     & {\color[HTML]{F8A102} \textbf{-0.5115}} & {\color[HTML]{F8A102} \textbf{1.8*1e-18}} & {\color[HTML]{C0C0C0} 0.0423}  & {\color[HTML]{C0C0C0} 0.4999} \\
\multirow{-4}{*}{\textbf{\begin{tabular}[c]{@{}c@{}}OHT3\\ MLP\end{tabular}}} & \multirow{-2}{*}{\textbf{Adam}} & L2                     & 0.5582         & 2.3*1e-22        & -0.7026          & 2.1*1e-39          & \multirow{-4}{*}{\textbf{\begin{tabular}[c]{@{}c@{}}Image\\ ResNet\end{tabular}}} & \multirow{-2}{*}{\textbf{Adam}} & L2                     & {\color[HTML]{F8A102} \textbf{-0.2876}} & {\color[HTML]{F8A102} \textbf{2.8*1e-06}} & {\color[HTML]{C0C0C0} 0.0197}  & {\color[HTML]{C0C0C0} 0.7538} \\ \hline
\end{tabular}

    }
    \label{app:tab:rho_p}
    \vskip -0.05in
\end{table*}

\end{document}